\pdfoutput=1
\documentclass{article}
\usepackage{spconf,amsmath,graphicx}

\usepackage{enumitem}
\setlist{nosep, leftmargin=14pt}

\usepackage{mwe} 

\usepackage{xcolor}

\usepackage{multirow}
\usepackage{stfloats}
\usepackage{lscape}

\usepackage[backref]{hyperref}

\title{Hybrid Spiking Neural Networks Fine-tuning for Hippocampus
  Segmentation}

\name{Ye Yue$^1$, Marc Baltes$^1$, Nidal Abujahar$^1$, Tao Sun$^1$,
  Charles D. Smith$^2$, Trevor Bihl$^3$, Jundong
  Liu$^1$\thanks{Corresponding author: Dr. Jundong Liu. Email:
    liuj1@ohio.edu.
  } } \address{ $^1$School of Electrical Engineering and Computer
  Science, Ohio University \\ $^2$ Department of Neurology, University
  of Kentucky \\ $^3$Department of Biomedical, Industrial $\&$ Human
  Factors Engineering \\  Wright State University}

\begin{document}
%
    \maketitle
    \begin{abstract}



Over the past decade, artificial neural networks (ANNs) have made
tremendous advances, in part due to the increased availability of
annotated data. However, ANNs typically require significant power and
memory consumptions to reach their full potential. Spiking neural
networks (SNNs) have recently emerged as a low-power alternative to
ANNs due to their sparsity nature.
  
SNN, however, are not as easy to train as ANNs. In this work, we
propose a hybrid SNN training scheme and apply it to segment human
hippocampi from magnetic resonance images. Our approach takes ANN-SNN
conversion as an initialization step and relies on spike-based
backpropagation to fine-tune the network.
Compared with the conversion and direct training solutions, our method
has advantages in both segmentation accuracy and training
efficiency. Experiments demonstrate the effectiveness of our model in
achieving the design goals.

\end{abstract}

    \begin{keywords}
        Spiking neural network, image segmentation, hippocampus,
        brain, U-Net, ANN-SNN conversion
    \end{keywords}

\section{Introduction}
\label{sec:intro}

Artificial Neural Networks (ANNs) have revolutionized many AI-related
areas, producing state-of-the-art results for a variety of tasks in
computer vision and medical image analysis.  
The remarkable performance of ANNs, however, often comes with a huge
computational burden, which limits their applications in power-hungry
systems such as edge and portable devices.
Bio-inspired spiking neural networks (SNNs), whose neurons imitate the
temporal and sparse spiking nature of biological neurons
\cite{roy2019towards,davies2021advancing, vicente2022keys,
  manna2022simple}, have recently emerged as a low-power alternative
for ANNs. SNN neurons process information with temporal binary spikes,
leading to sparser activations and natural reductions in power
consumption.

%
%

An SNN can be obtained by either converting from a fully trained ANN,
or through a direct training (training from scratch) procedure, where
a surrogate gradient is needed for the network to conduct
backpropagation.  Most ANN-SNN conversion solutions
\cite{diehl2015_conversion,rueckauer2016theory, snn-max2019,ho2021tcl}
focus on setting proper firing thresholds after copying the weights
from a trained ANN model. The converted SNNs commonly require a large
number of time steps to achieve comparable
accuracy, reducing the gains in power savings.
\cite{rathi_fine_tuning2020}.  Direct training solutions, on the other
hand, often suffer from expensive computation burdens on complex
network architectures \cite{shrestha2018slayer, wu2018spatio,
  rathi_fine_tuning2020,li2021differentiable}. For many pre-trained
ANNs on large datasets, e.g., ImageNet or LibriSpeech, training
equivalent SNNs from scratch would be very difficult.

Furthermore, most existing SNN works focus on
recognition related tasks.
Image segmentation, a very important task in medical
image analysis, is rarely studied, with the exception of
\cite{kim2021beyond, patel2021spiking}.
In \cite{kim2021beyond}, Kim {\it et al.} take a direct training
approach, which inevitably suffers from the common drawbacks of this
category.  Patel {\it et al.} \cite{patel2021spiking} use leaky {\it
  integrate-and-fire} (LIF) neurons for both ANNs and SNNs. While
convenient for conversion, the ANN networks are limited to a specific
type of activation functions and must be trained from scratch.

In this paper, we propose a hybrid SNN training scheme and apply it to
segment the human hippocampus from magnetic resonance (MR) images.
We use an ANN-SNN conversion step to initialize the weights and layer
thresholds in an SNN, and then apply a spike-based fine-tuning process
to adjust the network weights in dealing with potentially suboptimal
thresholds set by the conversion.
Compared with conversion-only and direct training methods, our
approach can significantly improve segmentation accuracy, as well as
decreases the training effort for convergence.
%
%

We choose the hippocampus as the target brain structure as accurate
segmentation of hippocampus provides a quantitative foundation for
many other analyses \cite{hobbs2016quad_short, shi2017nonlinear}, and
therefore has long been an important task in neuro-image research. A
modified U-Net \cite{ronneberger2015u} is used as the baseline ANN
model in our work.  To the best of our knowledge, this is the first
hybrid SNN fine-tuning work proposed for the image segmentation task,
as well as on U-shaped networks.

\section{Background}

\subsection{Hippocampus segmentation}
%

Segmentation of brain structures from MR images is an important task
in many neuroimage studies because it often in- fluences the outcomes
of other analysis steps. Among the anatomical structures to be
delineated, hippocampus is of particular interest, as it plays a
curial role in memory formation. It is also one of the brain
structures to suffer tissue damage in Alzheimer’s Disease. Traditional
solutions for automatic hippocampal segmentation include atlas-based
and patch-based methods
\cite{coupe2011patch,tong2013segmentation,song2015progressive}, which
commonly rely on identifying certain similarities between the target
image and the anatomical atlas, to infer labels for individual voxels.

In recent years, deep learning models, especially U-net
\cite{ronneberger2015u} and its variants, have become the dominant
solutions for medical image segmentation.  We have developed two
network-based solutions for hippocampus segmentation
\cite{chen2017accurate,chen2017hippocampus}, producing
state-of-the-art results. In \cite{chen2017accurate}, we proposed a
multi-view ensemble convolutional neural network (CNN) framework in
which multiple decision maps generated along different 2D views are
integrated. In \cite{chen2017hippocampus}, an end-to-end deep learning
architecture is developed that combines CNN and recurrent neural
network (RNN) to better leverage the dimensional anisotropism in
volumetric medical data.




\subsection{Spiking neural network optimization}
%

Gradient descent-based backpropagation is by far the most used method
to train ANN models. Unfortunately, the neurons in SNNs are highly
non-differentiable with a large temporal aspect,
which makes gradient descent not directly applicable to training SNNs.

The direct training approach works around this issue through surrogate
gradients \cite{wu2018spatio, kim2020revisiting}, which are
approximations of the step function that allow the backpropagation
algorithm to be conducted to update the network weights. In terms of
assigning spatial and temporal gradients along neurons,
spike-timing-dependent plasticity (STDP) is a popular solution, which
actively adjusts connection weights based on the firing timing of
associated neurons \cite{liu2020unsupervised}.

Training an ANN first and converting it into an SNN can completely
circumvent the non-differentiability problem. 
One major group of conversion solutions
\cite{diehl2015_conversion,rueckauer2016theory, snn-max2019,ho2021tcl}
train ANNs with rectified linear unit (ReLU) neurons and then convert
them to SNNs with IF neurons by setting appropriate firing thresholds.
Hunsberger and Eliasmith \cite{hunsberger2015spiking,
  rasmussen2019nengodl} use soft LIF neurons, which have smoothing
operations applied around the firing threshold. As a result of the
smoothing, gradient-based backpropagation can be carried out to train
the network. This design makes the conversion from ANN to SNN rather
straightforward.

\section{Method}

\begin{figure}[htb]
    \centering
     \centerline{\includegraphics[width=0.45\textwidth]{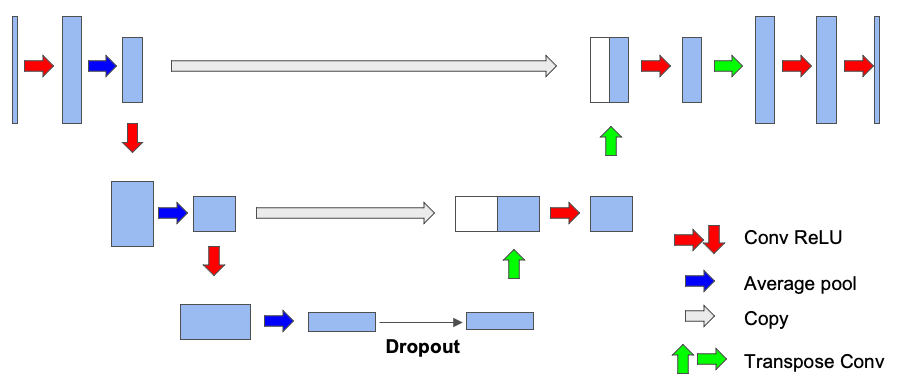}}
    \caption{Network architecture of our baseline ANN model. }
    \label{fig:unet_structure}
\end{figure}

{\bf ANN baseline model} We use a modified U-Net as the baseline ANN
model in this work, which also follows an encoding and decoding
architecture, as shown in Fig.~\ref{fig:unet_structure}. Taking 2D
images as inputs, the encoding part repeats the conventional
convolution + pooling layers to extract high-level latent
features. The decoding part reconstructs the segmentation ground-truth
mask by using transpose / deconvolution layers.  To fully exploit the
local information,
{\it skip connections} are introduced to concatenate
the corresponding features maps between encoding stack and decoding
stack.

Due to the constraints imposed by ANN-SNN conversion, a number of
modifications have to be made from the original U-Net, as well as our
previous hippocampus segmentation networks \cite{chen2017accurate,
  chen2017hippocampus}. %
%
First, we replace max-pooling with average-pooling, as there is no
effective implementation of max-pooling in SNNs. Second,
the bias components of the neurons are removed, as
they may interfere with the voltage thresholds in SNNs,
complicating the training of the networks.  Third, batch normalizations
are removed, due to the absence of the bias terms. Last, {\it dropout
  layers} are added to provide regularization for both ANN and SNN
training.

\subsection{Our proposed hybrid SNN fine-tuning}

Inspired by a previous work on image classification network
\cite{rathi_fine_tuning2020}, we develop a hybrid SNN fine-tuning
scheme for our segmentation network. We first train an ANN to its full
convergence and then convert it to a spiking network with reduced time
steps.  The converted SNN is then taken as an initial state for a
fine-tuning procedure.


%
%

{\bf The SNN neuron} model in our work is integrate-and-fire (IF)
model where the membrane potential will not decrease during the time
when neuron does not fire as opposed to LIF 
neurons.
The dynamics of our IF neurons
can be described as:

\begin{equation}
  u_{i}^t = u_{i}^{t-1} + \sum_j w_{ij}o_j - v o_i^{t-1}
  \label{eqn:IF}
\end{equation}

\begin{equation}
  o_i^{t-1} =
  \begin{cases}
    1 & \textrm{if } 
    u_i^{t-1} > v \\
    0   & \textrm{otherwise}
  \end{cases}
  \label{eqn:firing}
\end{equation}
where $u$ is the membrane potential,
$t$ is the time step, subscript $i$ and $j$ represent the post- and
pre-neuron, respectively,$w$ is the weight connecting the pre- and
post-neuron, $o$ is the binary output spike, and $v$ is the firing
threshold. Each neuron integrates the inputs from the previous layer
into its membrane potential, and reduces the potential by a threshold
voltage if a spike is fired.
%
%

Our SNN network has the same architecture as the baseline ANN, where
the signals transmitted within the SNN are rate-coded spike trains
generated through a Poisson generator.  During the conversion process,
we load the weights of the trained ANN into the SNN network and set
the thresholds for all layers to 1s. Then a threshold balancing
procedure \cite{snn-max2019} is carried out to update the threshold of
each layer.


{\bf Fine-tuning} of the converted SNN is conducted using spike-based
backpropagation. It starts at the output layer, where the signals are
continuous membrane potentials, generated through the summation:
\begin{equation}
  u_{i}^t = u_{i}^{t-1} + \sum_j w_{ij}o_j
  \label{eqn:o_layer}
\end{equation}
The number of neurons in the output layer is the same as the size of
the input image. Compared with the hidden layer neurons in
Eqn.~\ref{eqn:IF}, the output layer does not fire and therefore the
voltage reduction term is removed. Each neuron in the output layer is
connected to a Sigmoid activation function to produce the predictive
probability of the corresponding pixel belonging to the target area
(hippocampus).
%


Let $L(\cdot)$ be the loss function defined based on the ground-truth
mask and the predictions. In the output layer, neurons do not generate
spikes and thus do not have the non-differentiable problem.  The
update of the hidden layer parameters $W_{ij}$ is described by:

\begin{equation}
  \Delta W_{ij} = \sum_t \frac{\partial L}{\partial W_{ij}^t} = \sum_t
  \frac{\partial L}{\partial o_{i}^t} \frac{\partial o_{i}^t}{\partial
    u_i^t}\frac{\partial u_i^t}{\partial W_{ij}^t}
  \label{eqn:delta}
\end{equation}

Due to the non-differentiability of spikes, a surrogate gradient-based
method is used in backpropagation. In \cite{rathi_fine_tuning2020},
the authors propose a surrogate gradient function $\frac{\partial
  o^t}{\partial u^t} = \alpha e^{-\beta \Delta t}$. In this work,
we choose a linear approximation proposed in
\cite{bellec2018long}, which is described as:
 \begin{equation}
    \frac{\partial o^t}{\partial u^t} = \alpha \max \{0, 1-\left| u^t -V_t
    \right|\}\label {eq:Linear_Gradient}
 \end{equation}
 where $V_t$ is the threshold potential at time $t$, and
 $\alpha$ is a constant.
%

Our {\it conversion + fine-tuning} follows the same framework proposed
in \cite{rathi_fine_tuning2020}, which demonstrates that such hybrid
approach can achieve, with much fewer time steps, similar accuracy
compared to purely converted SNNs, as well as faster convergence than
direct training methods. It should be noted that our work is the first
attempt of exploring the application of spike-based fine-tuning and
threshold balancing on fully convolutional networks (FCNs), including
the U-Net.

    \subsection{Different losses}

We explore different loss functions in this work, which include
{\it binary cross entropy} (BCE), {\it Dice loss} and a combination of
the two losses (BCE-Dice).  BCE loss is the average of per-pixel loss
and gives an unbiased measurement of pixel-wise similarity between
prediction and ground-truth: 
\begin{equation}
    L_{\textrm{BCE}} = \sum_{i=1}^{N}-[r_i\log s_i +(1-r_i)\log(1-s_i)]
\end{equation}
where $s_i \in [0,1]$ is the predicted value of a pixel and $r_i \in
\{0,1\}$ is the ground-truth label for the same pixel.
$\epsilon$ is a small number to smooth the loss,
which is set to $10^{-5}$ in our experiments.
{\it Dice loss}
focuses more on the extent to which the predicted and ground-truth
overlap:
\begin{equation}
  L_\textrm{Dice} = \frac{2\sum_i s_{i}r_{i} + \epsilon}{\sum_i
    s_{i}+\sum_i r_{i} + \epsilon}
\end{equation}

We also explore the effects of a weighted combination of BCE and Dice:
$L_\textrm{BCE\_Dice} = 0.3 \times L_\textrm{BCE} + 0.7 \times
L_\textrm{Dice}$


\section{Experiments and Results}

{\bf The data} used in this work were obtained from the ADNI database
(\url{https://adni.loni.usc.edu/}).  In total, 110 baseline
T1-weighted whole brain MRI images from different subjects along with
their hippocampus masks were downloaded.  In our experiments we only
included normal control subjects. Due to the fact that the hippocampus
only occupies a very small part of the whole brain and its position in
the whole brain is relatively fixed, we roughly cropped the right
hippocampus of each subject and use them as the input for
segmentation. The size of the cropping box is $24 \times 56 \times
48$.


\subsection{Training and testing}
We train and evaluate our proposed model using a 5-fold
cross-validation, with 22 and 88 subjects in the test and training
sets, respectively, in each fold.  The batch size in training and
testing of ANN and SNN is set to 26.  Training of both ANN and the SNN
networks uses the Adam optimizer with slightly different
parameters. The learning rate for training both networks is initially
set to 0.001 and is later adjusted by the ReduceLROnPlateau scheduler
in PyTorch, which monitors the loss during training, and tunes down
the learning rate when it finds the loss stops dropping.  Following a
similar setup in \cite{rathi_fine_tuning2020}, we use time steps of
200 for the ANN-SNN conversion routine.  The ANN models were trained
with 100 epochs and SNN models were trained over 35 epochs.  Also, we
repeat the ANN $\rightarrow$ Conversion $\rightarrow$ Fine-tuning
procedure with three different loss functions: BCE only, Dice only and
combined BCE and Dice.

\subsection{Results}

In this section, we present and evaluate the experimental results for
the proposed model. Two different performance metrics, 3D Dice ratio
and 2D slice-wise Dice ratio, were used to measure the accuracy of the
segmentation models. The 3D Dice ratio was calculated subject-wise for
each 3D volume. Mean and standard deviation averaged from 5 folds are
reported. The 2D slice based Dice ratio was calculated slice by slice,
and the mean and standard deviation were averaged from all test
subjects’ slices.

Accuracies of the model on the test data are summarized in Table
\ref{tab:table_exp_result}. The best performance for the ANN and
fine-tuned SNN are highlighted with bold font. It is evident that
network accuracies drop significantly after conversion (middle column)
and our fine-tuning procedure can bring the performance of SNNs (right
column) back close to the ANN level. The models built on the three
loss functions have comparable performance in ANNs and fine-tuned
SNNs, with Dice loss has slight edge over BCE and BCE-Dice combined.


Most fine-tuning procedures converge much faster than the maximal 35
epochs. On average, the three models in our experiments take 10.11
epochs to converge (with a standard deviation of 4.90).  This achieves
a great reduction in training complexity compared to the direct
training models, which take roughly 60 epochs on average.


\begin{table*}
    \centering
    \scalebox{0.95}{
        \begin{tabular}{|c|cc|cc|cc|}
            \hline
            \multirow{2}{*}{{Loss }} &
            \multicolumn{2}{c|}{{ANN Accuracy}} &
            \multicolumn{2}{c|}{{Converted-SNN Accuracy}} &
            \multicolumn{2}{c|}{{Fine-tuned SNN Accuracy}} \\ \cline{2-7}
            &
            \multicolumn{1}{c|}{2D} &
            3D &
            \multicolumn{1}{c|}{2D} &
            3D &
            \multicolumn{1}{c|}{2D} &
            3D \\ \hline \hline
            {BCE} &
            \multicolumn{1}{c|}{$77.76 \pm 2.40$} &
            $83.17 \pm 1.44$ &
            \multicolumn{1}{c|}{$30.51 \pm 9.93$} &
            $21.60 \pm 12.39$ &
            \multicolumn{1}{c|}{$76.92 \pm 3.77$} &
            $81.83 \pm 2.99$ \\ \hline
            {Dice} &
            \multicolumn{1}{c|}{${\bf 78.86 \pm 2.39}$} &
            ${\bf 84.21 \pm 1.66}$ &
            \multicolumn{1}{c|}{$61.78 \pm 3.32$} &
            $65.90 \pm 6.94$ &
            \multicolumn{1}{c|}{${\bf 78.09 \pm 3.85}$} &
            $81.86 \pm 4.40$ \\ \hline
            {BCE-Dice} &
            \multicolumn{1}{c|}{$78.58 \pm 2.44$} &
            $83.14 \pm 1.59$ &
            \multicolumn{1}{c|}{$52.03 \pm 9.91$} &
            $52.78 \pm 14.73$ &
            \multicolumn{1}{c|}{$77.72 \pm 3.69$} &
            ${\bf 81.95 \pm 3.24}$ \\ \hline
        \end{tabular}%
    }
    \caption{Average accuracies of ANNs, converted SNNs and fine-tuned
      SNNs built on three difference loss functions.  }
    \label{tab:table_exp_result}
\end{table*}

%
%
%

\begin{figure}[htb]

\begin{minipage}[b]{0.48\linewidth}
  \centering
  \centerline{\includegraphics[width=3.4cm]{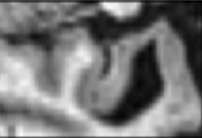}}
  \centerline{(a) Input slice}\medskip
\end{minipage}
\hfill
\begin{minipage}[b]{.48\linewidth}
  \centering
  \centerline{\includegraphics[width=3.6cm]{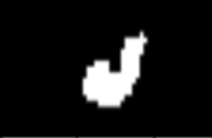}}
  \centerline{(b) Ground-truth mask}\medskip
\end{minipage}

\begin{minipage}[b]{0.48\linewidth}
  \centering
  \centerline{\includegraphics[width=3.4cm]{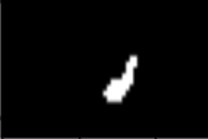}}
  \centerline{(c) Converted SNN}\medskip
\end{minipage}
\hfill
\begin{minipage}[b]{0.48\linewidth}
  \centering
  \centerline{\includegraphics[width=3.6cm]{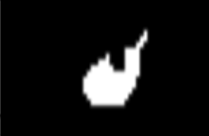}}
  \centerline{(d) Fine-tuned SNN}\medskip
\end{minipage}
\caption{An example slice of (a) input; (b) ground-truth mask; (c)
  segmentation result from the converted SNN; and (d) result from the
  fine-tuned SNN.} 
\label{fig:res}
\end{figure}

\begin{figure}[htb]

    \centering
    \centerline{\includegraphics[width=7.5cm]{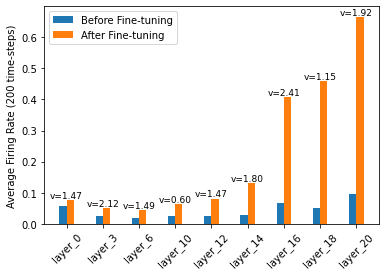}}
    \caption{Firing frequencies of neurons in different layers. Blue
      bars show those for a converted-SNN and orange bars are for the 
      fine-tuned SNN. $v$ is the layer threshold. }
    \label{fig:firing_rate}
\end{figure}

In order to find out how the fine-tuning procedure improves the
segmentation accuracy, we look into details of both the outputs and
the internal spiking patterns of the networks.  Fig.~\ref{fig:res}
shows an example of input slice, ground-truth mask and the
corresponding outputs from the converted and fine-tuned SNNs. We can
see the output mask from the converted SNN (Fig.~\ref{fig:res}.(c)),
while carrying a similar shape, is much smaller than the
ground-truth. We believe the reason is
that many neurons are not sufficiently activated due to the suboptimal
thresholds set by the conversion procedure. The proposed fine-tuning
step, on the other hand, can update the network weights to adjust for
such thresholds, bringing the neurons back to active for improved
accuracy. To confirm this thought, we record the firing frequency of
each layer in the SNN models before and after the fine-tuning and plot
them in Fig.~\ref{fig:firing_rate}. It is evident that neurons become
more active after the fine-tuning, producing more accurate
segmentation predictions.





\section{Conclusions}

In this work, we present a hybrid ANN-SNN fine-tuning scheme for the
image segmentation problem. Our approach is rather general and can
potentially be applied to many applications based on U-shaped
networks. We take the human hippocampus as the target structure, and
demonstrate the effectiveness of our approach through experiments on
ADNI data. Exploring more applications is among our next steps.

%
%
%

    \small
\bibliographystyle{IEEEbib}
\bibliography{snn,seg,my_papers}
\end{document}